\pdfoutput=1

\documentclass[11pt]{article}

\usepackage{ACL2023}
\usepackage{times}
\usepackage{latexsym}
\usepackage{booktabs}       
\usepackage{xcolor}
\usepackage{marvosym}
\usepackage{times}
\usepackage{latexsym}
\usepackage{graphicx}
\usepackage{roboto}  
\usepackage[T1]{fontenc}

\usepackage[utf8]{inputenc}
\usepackage{url}            
\usepackage{booktabs}       
\usepackage{amsfonts}       
\usepackage{nicefrac}       
\usepackage{microtype}      
\usepackage{xcolor}         
\usepackage{graphicx}         
\usepackage{epsfig}
\usepackage{amsmath}
\usepackage{multirow}
\usepackage{array}
\usepackage{pifont}
\usepackage{booktabs}
\usepackage{wrapfig}
\usepackage{hhline}
\usepackage{mathrsfs} 
\usepackage{algorithmic}
\usepackage[vlined,ruled]{algorithm2e}
\usepackage{color, colortbl}
\usepackage{bm}
\usepackage{tikz}
\usepackage{marvosym}
\usepackage{epigraph}
\usepackage{ifsym}
\usepackage{mleftright}
\usepackage{icomma}
\usepackage{xspace} 
\usepackage{fontawesome}
\usepackage{float}

\newcommand{\eg}{e.g.\xspace}
\newcommand{\ie}{i.e.\xspace}
\usepackage{microtype}
\usepackage{diagbox} 
\usepackage{threeparttable}
\usepackage{subcaption}
\usepackage{inconsolata}

\definecolor{steelBlue}{RGB}{43, 129, 214}
\definecolor{savoyBlue}{RGB}{86, 100, 191}
\definecolor{royalPurple}{RGB}{129, 71, 169}
\definecolor{fandango}{RGB}{171, 41, 146}
\definecolor{magentaDye}{RGB}{214, 12, 123}
\definecolor{Gray}{gray}{0.85}
\definecolor{LightCyan}{rgb}{0.88,1,1}
\definecolor{steelBlue}{RGB}{43, 129, 214}
\definecolor{savoyBlue}{RGB}{86, 100, 191}
\definecolor{royalPurple}{RGB}{129, 71, 169}
\definecolor{fandango}{RGB}{171, 41, 146}
\definecolor{magentaDye}{RGB}{214, 12, 123}

\title{\textcolor{steelBlue}{M}\textcolor{savoyBlue}{M}\textcolor{royalPurple}{I}\textcolor{fandango}{n}\textcolor{magentaDye}{A}: Benchmarking \textcolor{steelBlue}{M}ultihop \textcolor{savoyBlue}{M}ultimodal \textcolor{royalPurple}{I}\textcolor{fandango}{n}ternet \textcolor{magentaDye}{A}gents}

\usepackage{pifont}%
\newcommand{\cmark}{\textcolor[rgb]{0.13, 0.55, 0.13}{\ding{51}}}%
\newcommand{\xmark}{\textcolor[rgb]{0.5, 0, 0}{\ding{55}}}%

\usepackage[accsupp]{axessibility}  
\usepackage{soul}

\makeatletter
\renewcommand\paragraph{
  \@startsection{paragraph} 
  {4} 
  {\z@} 
  {.5em \@plus1ex \@minus.2ex} 
  {-1.5em} 
  {\normalfont\normalsize\bfseries} 
}
\makeatother

\title{MMInA: Benchmarking Multihop Multimodal Internet Agents}

 \author{
 Shulin Tian$^{\ast}$~
 Ziniu Zhang$^{\ast}$ ~ 
  Liangyu Chen$^{\ast,\dagger}$~
  \textbf{Ziwei Liu \Letter}~
 \\
 S-Lab, Nanyang Technological University
 \\
  \texttt{\{shulin002, lchen025, ziwei.liu\}@ntu.edu.sg}  \\
  \texttt{michaelzhangziniu@gmail.com} \\
}

\begin{document}
\maketitle
\footnotetext{$\ast$~Equal Contribution. $\quad$ $\dagger$~Project Lead.$\quad$ \Letter~Corresponding Author.}

\begin{abstract}

Autonomous embodied agents live on an Internet of multimedia websites. Can they hop around multimodal websites to complete complex user tasks? Existing benchmarks fail to assess them in a realistic, evolving environment for their embodiment across websites. To answer this question, we present \textbf{MMInA}, a multihop and multimodal benchmark to evaluate the embodied agents for compositional Internet tasks, with several appealing properties: 
\textbf{1)} \textit{Evolving real-world multimodal websites.} Our benchmark uniquely operates on evolving real-world websites, ensuring a high degree of realism and applicability to natural user tasks. Our data includes 1,050 human-written tasks covering various domains such as shopping and travel, with each task requiring the agent to extract multimodal information from web pages as observations autonomously;
\textbf{2)} \textit{Multihop web browsing.} Our dataset features naturally compositional tasks that require information from or actions on multiple websites to solve, to assess long-range reasoning capabilities on web tasks; 
\textbf{3)} \textit{Holistic evaluation.} We propose a novel protocol for evaluating an agent's progress in completing multihop tasks. We experiment with both standalone (multimodal) language models and heuristic-based web agents.
Extensive experiments demonstrate that while long-chain multihop web tasks are easy for humans, they remain challenging for state-of-the-art web agents.
We identify that agents are more likely to fail on the early hops when solving tasks with more hops, which results in lower task success rates.
To address this issue, we propose a simple memory augmentation approach that replays past action trajectories to reflect. Our method significantly improves the performance of both the single-hop and multihop web browsing abilities. Our code and data are available on \href{https://github.com/shulin16/MMInA}{github.com/shulin16/MMInA}.

\end{abstract}
\section{Introduction}
\label{sec:intro}

Building embodied agents capable of autonomous behaviors navigating in various environments has been a longstanding and intricate challenge in the realm of artificial intelligence research~\cite{maes1993modeling,ziemke1998adaptive,florian2003autonomous, steels2018artificial}. One common scenario that necessitates automation involves the interaction with digital interfaces~\cite{puig2018virtualhome,toyama2021androidenv}, with a particular emphasis on the automation of actions performed on rich Internet websites~\cite{shi2017world,yao2023react,hong2024cogagent}. Real-world web tasks are inherently compositional, requiring multihop actions across multiple websites. To accomplish this, agents must possess both \textit{long-range planning} and \textit{multimodal reasoning} capabilities. This includes understanding high-level instructions from user inputs, planning multihop actions across the web browser environment by leveraging HTML content and visual cues, and making informed predictions based on observations. 
While web agents perform well on single-hop tasks that require interactions with only one website, according to existing benchmarks~\cite{li2023api, liu2024agentbench, liu2023bolaa, zhou2024webarena, koh2024visualwebarena}, we observe that most agents struggle with multihop web tasks, which are prevalent in real-world scenarios where users must gather information from or take actions across multiple websites to accomplish a high-level task
(Tab.~\ref{tab:results}). This gap motivates us to establish a multihop web browsing benchmark to assess the usefulness of Internet agents in natural multihop tasks.

\begin{figure*}[t]
\centering
    \hspace{-7mm}\includegraphics[width=1.05\textwidth]{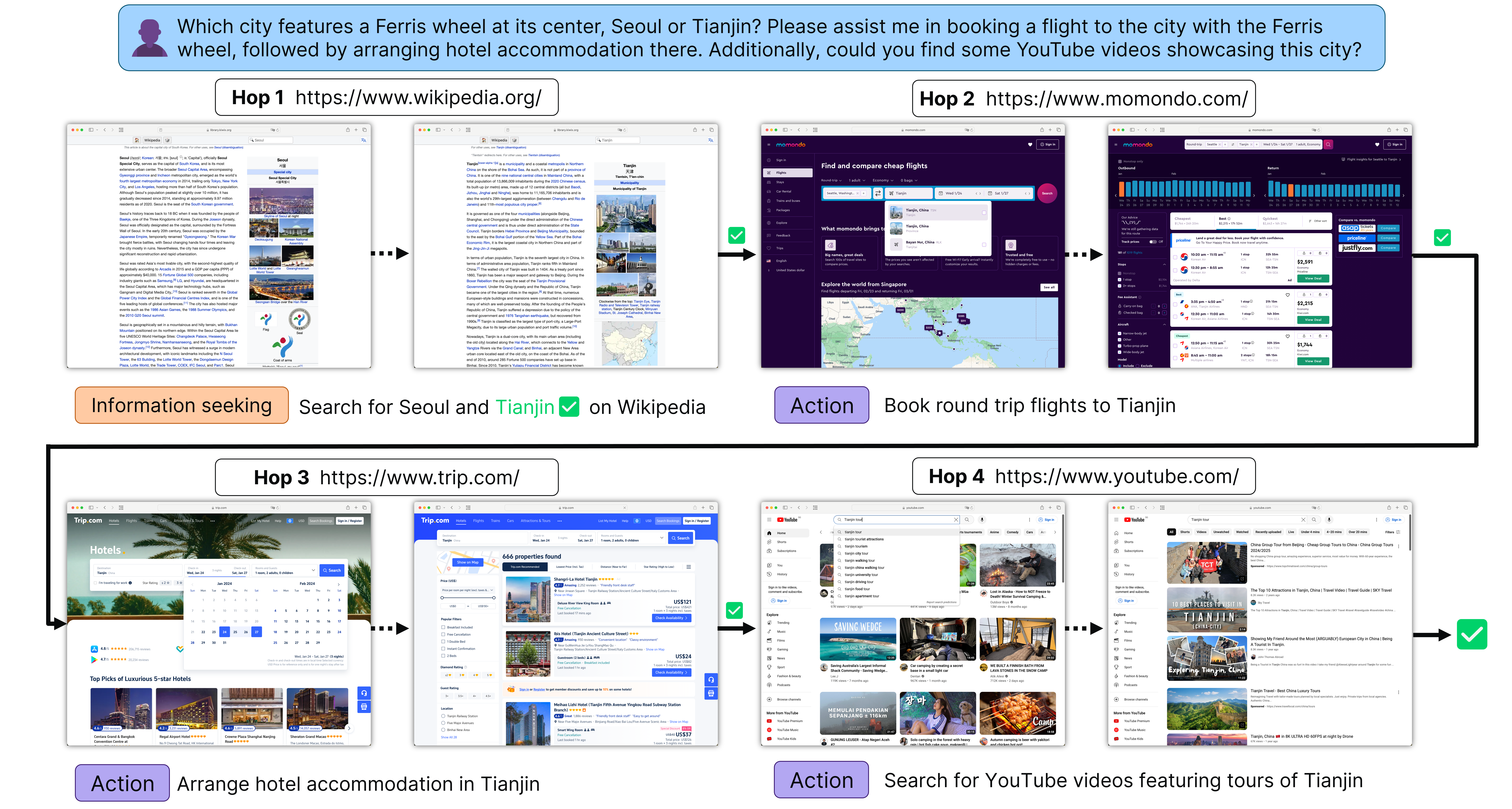}
    \captionof{figure}{\textbf{An example task from MMInA}.
    To evaluate an Internet agent's ability to carry out complex tasks, we make it navigate through a variety of websites to gather information and execute actions. In our proposed holistic evaluation protocol, each phase of the compositional task (defined as a \textit{hop}) and the overall task are assessed for performance. Our benchmark includes 1,050 varied human-written multimodal tasks that require an average of 2.85 hops between websites and 12.9 actions to complete. The longest compositional task takes 10 hops.}
    \vspace{-10pt}
    \label{fig:teaser}
\end{figure*}

Another gap in web agent research is multimodality. Existing benchmarks pose autonomous agent tasks that rely solely on textual information \cite{zhou2024webarena,deng2023mind2web,yao2023react}. However, in real-world scenarios, visual information often plays an indispensable role and cannot be disregarded. For example, consider the task of ``Help me purchase a blue cotton shirt'', where the color attribute, derived from visual information, becomes crucial in fulfilling the user's request. However, there is a notable lack of current web-based agent benchmarks with emphasis on assessing the capabilities of comprehending and interacting with both textual and visual inputs, where a significant number of them primarily concentrate on tasks that involve text-based interactions.

To address the two issues above, we present MMInA, a novel benchmark designed to advance multihop and multimodal Internet task solving. 
MMInA operates on real-world, evolving websites (Fig.~\ref{fig:teaser}), offering high realism and applicability to natural scenarios (Tab.~\ref{tab:vwa_comparison}). It focuses on realistic tasks users commonly perform, such as navigating e-commerce platforms, extracting information from content-rich sites like Wikipedia, and conducting comparative analysis across web sources. Our 1,050 human-written tasks challenge agents to process multimodal inputs across multiple website hops and execute complex, multi-step reasoning, surpassing the simpler tasks found in existing benchmarks.

Our experiments with state-of-the-art models as agents' reasoning backbones show that while progress has been made in handling simple textual tasks, the integrated and sequential nature of tasks in MMInA remains a significant challenge. For example, GPT-4V~\cite{achiam2023gpt} achieves a 21.8\% success rate, which improves upon textual baselines but falls short of human performance (96.3\%). Agents often fail early in multi-hop tasks (Tab.~\ref{tab:tri}), leading to lower success rates. These results highlight the complexity of real-world web navigation and decision-making, underscoring the need for further advancements in multimodal and multihop reasoning. To bridge this gap, we propose a memory augmentation approach that replays past action trajectories, significantly improving both single-hop and multihop performance. This model-agnostic technique can be applied to other large multimodal models in the future.


In summary, our contributions are as follows:
\begin{itemize}
    \vspace{-6pt}
    \item We introduce MMInA, an Internet agent benchmark featuring 1,050 multihop multimodal tasks across 14 diverse, evolving websites with realistic features. The multihop tasks assess more complex, human-like problem-solving actions, closely resembling real-world scenarios. We evaluate current large language models (LLMs) and large multimodal models (LMMs) as agents' backbones on the benchmark.
    \vspace{-6pt}
    \item  We propose a holistic evaluation method for multihop tasks. Building on the limitations of task success rate evaluation, which often yields poor agent performance and limited insights, we propose a new protocol for multihop tasks based on hop success rate. This provides a more granular assessment, offering deeper insights into the relationship between agent behavior and hop length, facilitating a more informed analysis of long-chain reasoning capabilities. 
    \vspace{-6pt}
    \item We propose a lightweight and adaptable memory-augmented method that enhances agent performance by leveraging past action histories.

\end{itemize}

\begin{table*}[t]
\centering
\caption{\textbf{Comparison between the MMInA benchmark and related benchmarks.} MMInA employs a flexible environment that supports agents to generate open-ended actions. We selected 14 evolving real-world websites to benchmark multihop multimodal Internet agents, which can be easily expanded for future deployments.}
\resizebox{\textwidth}{!}{%
\begin{tabular}{c|c|c|c|c|c}
\toprule
\textbf{Benchmark} & \textbf{Multi-modal} & \textbf{Max / Avg. Hops} & \textbf{Website Type} & \textbf{Dynamic Interaction} & \textbf{\# Websites} \\
\midrule
MiniWoB++~\cite{liu2018reinforcement} & \cmark & 1 / 1.00 & Static simplified websites & \cmark~(Open-ended) & 100 \\
WebShop~\cite{yao2022webshop} & \cmark & 1 / 1.00 & Static simplified websites & \cmark~(Open-ended) & 1 \\
Mind2Web~\cite{deng2023mind2web} & \xmark & 1 / 1.00 & Static real-world websites & \xmark~(MC) & 131 \\
RUSS~\cite{xu2021grounding} & \xmark & 2 / 1.10 & Static real-world websites & \xmark~(MC) & 22 \\
WebArena~\cite{zhou2024webarena} & \xmark & 2 / 1.06 & Static real-world websites & \cmark~(Open-ended) & 6 \\
VWA~\cite{koh2024visualwebarena} & \cmark & 2 / 1.05 & Static real-world websites & \cmark~(Open-ended) & 3 \\
WebVoyager~\cite{he2024webvoyager} & \cmark & 4 / 2.40 & Dynamic real-world websites & \cmark~(Open-ended) & 15 \\
\midrule
MMInA & \cmark & 10 / 2.85 & Evolving real-world websites & \cmark~(Open-ended) & 14 \\
\bottomrule
\end{tabular}%
}
\label{your-label}
\end{table*}

\section{Related Works}
\label{sec:related_works}

\paragraph{Agent Benchmarks and Environments}
Most existing works evaluate autonomous agents on curated textual I/O interfaces, leaving a gap in assessing their performance on real-world automation tasks. Tool-use benchmarks like~\cite{li2023api, patil2024gorilla,liu2024agentbench} aim to assess agents' performances with tool-usage capability; BOLAA~\cite{liu2023bolaa} is another benchmark that coordinates multiple autonomous agents to make collective decisions.
OpenAGI~\cite{ge2023openagi} and GAIA~\cite{mialon2024gaia} are multimodal benchmarks crafted for generalist agents that define multi-step tasks across multiple modalities.
However, none of the above benchmarks explored the usage of LLMs or LMMs in web browsing environments or posed an effective evaluation metric specifically tailored for web agent tasks.

\paragraph{Web Agents}

Webshop~\cite{yao2022webshop} builds a simulated e-commerce environment featuring 1.18 million real-world products, complemented by 12,087 crowdsourced textual instructions, while Mind2Web~\cite{deng2023mind2web}, CogAgent~\cite{hong2024cogagent}, and SeeAct~\cite{zheng2024gpt} try to construct a generalist web agent; WebVoyager~\cite{he2024webvoyager} can automatically identify interactive elements based on the types of webpage. Recently, WebArena~\cite{zhou2024webarena} deploys a standalone set of multicategory websites in an interactive environment, where VisualWebArena~\cite{koh2024visualwebarena} is a subsequent project that built upon WebArena, introducing the reliance on visual cues into the benchmark's design. However, we found that the tasks of existing benchmarks are oversimplified whose completions requiring a single website, which is highly diverged from the natural web browsing tasks and should originally be designed for multihop over a long-horizon setting.

\section{MMInA Benchmark}
\label{sec:theia}



\begin{figure*}[t]
\sbox0{
\begin{subfigure}[b]{\dimexpr 0.6\textwidth-0.5\columnsep}
  \includegraphics[width=\linewidth]{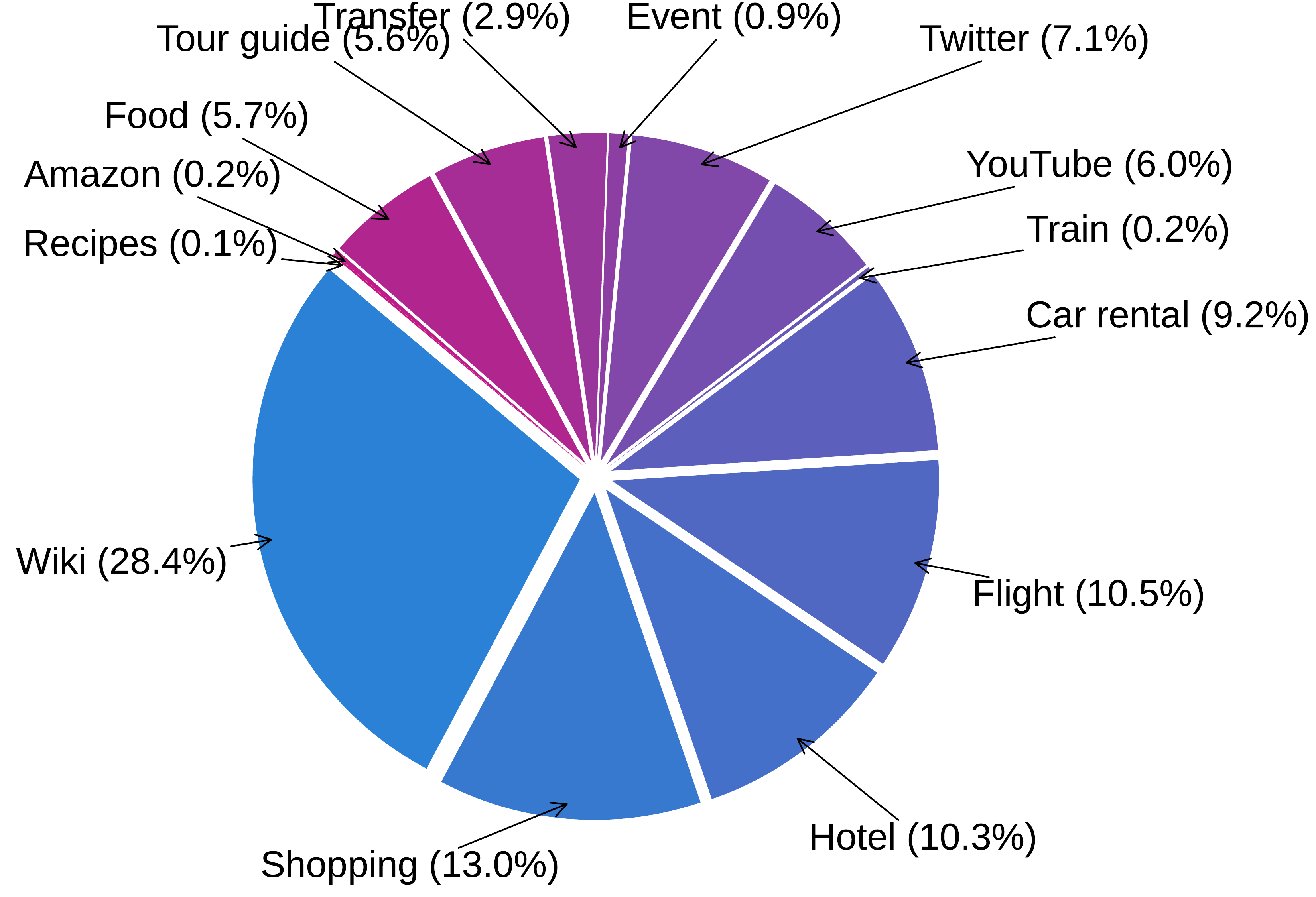}
  \caption{\textbf{Source websites of the 2,991 hops.}}
\end{subfigure}
}%
\usebox0\hfill\begin{minipage}[b][\ht0][s]{\wd0}
  \begin{subfigure}{\linewidth}
    \includegraphics[width=0.58\linewidth]{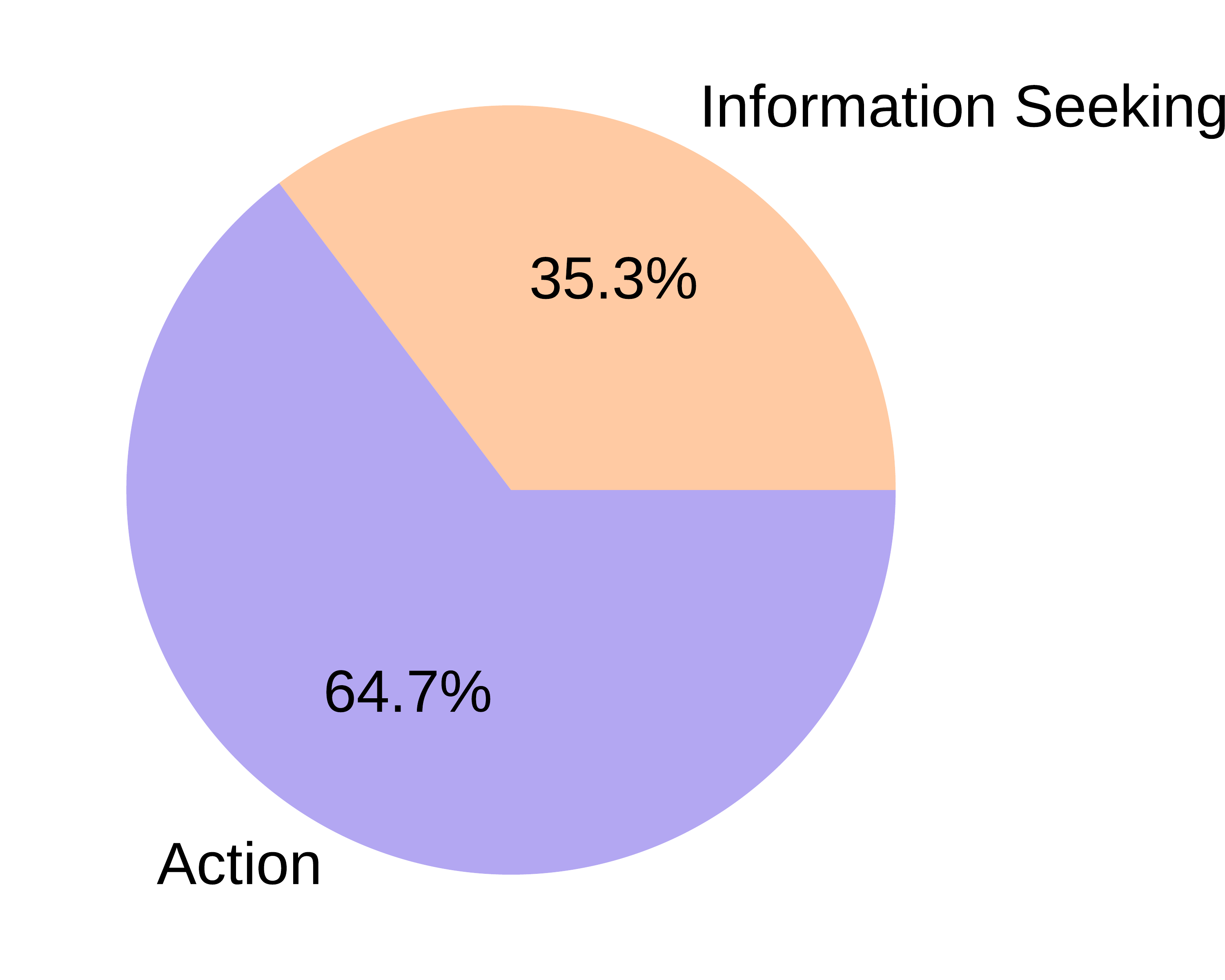}
    \vspace{-5mm}
    \caption{\textbf{Intent types of hops.}}
  \end{subfigure}\par
  \vfill
  \begin{subfigure}[b]{\linewidth}
    \includegraphics[width=0.58\linewidth]{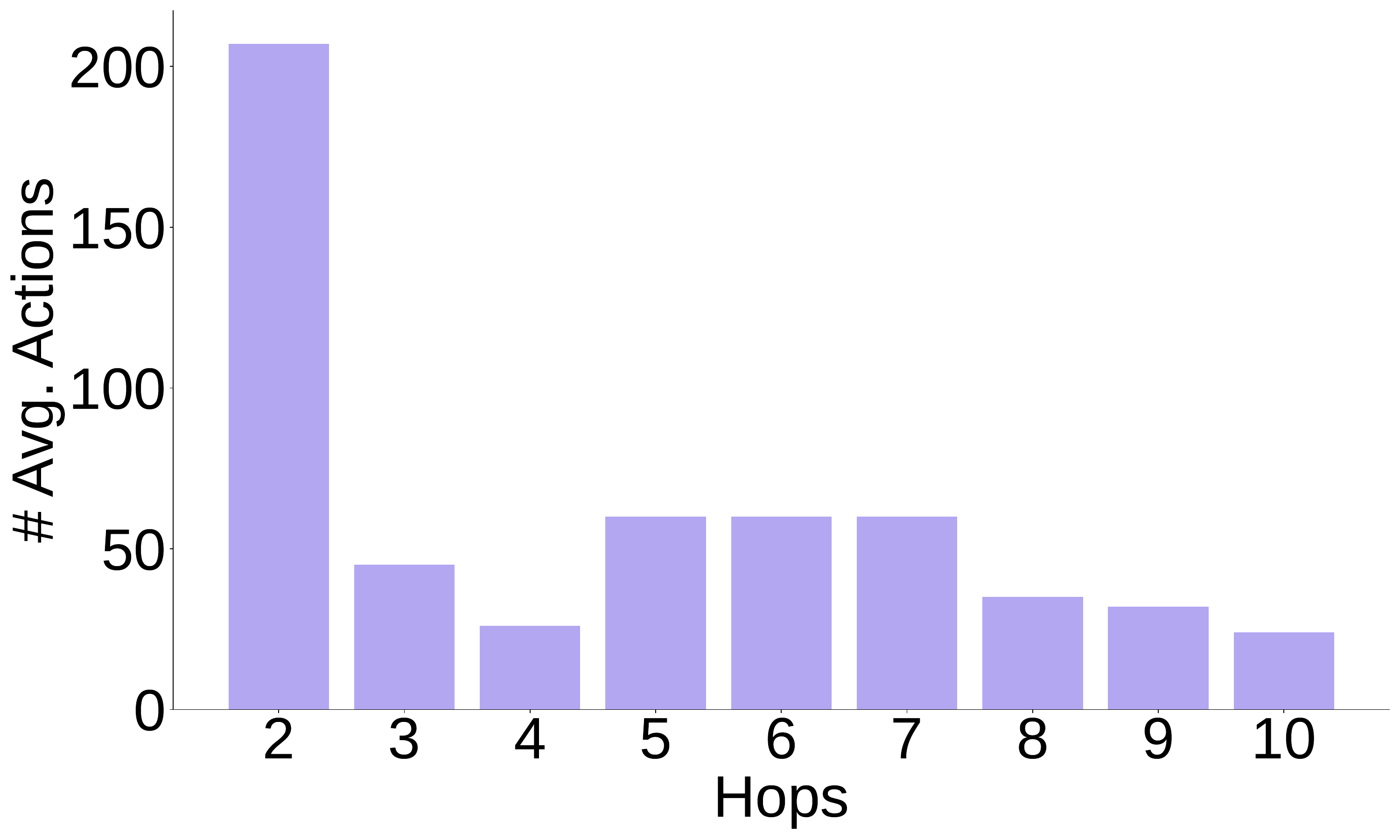}
    \caption{\textbf{Counts of multihop tasks.}}
  \end{subfigure}
\end{minipage}
\vspace{10mm}
\caption{\textbf{Statistics of the MMInA benchmark.} A web browsing task is composed of one or multiple hops between websites. MMInA covers diverse intents and domains of hops to resemble naturally compositional user tasks.}
\label{fig:stats}
\vspace{-5mm}
\end{figure*}




\subsection{Environment}

Following~\citet{zhou2024webarena}, we formulate web browsing as a partially observable Markov decision process $\langle S, A, P, R \rangle$. The state space $S$ is the whole Internet content, the status of the browser environment and agent, exceeding representable expressions in practice. Therefore, we pass the partial observation space $\Omega$ of $S$ to the agent. At each time step $t$, an agent arrives at a certain state of a particular web page. The accessibility tree of the screenshot with linked images, with the action/state histories, forms a partial observation $o_t \in \Omega$ for the agent. Then the web agent takes an action $a_t$ sampled from the action space $A$, being either an executable operation on the web page or a textual output as the answer to a question (Sec.~\ref{sec:evaluation}). The state transition probability matrix $P: S \times A \rightarrow S'$ is implicitly encoded as the world knowledge of the Internet environment, which can be inferred or learned by a web agent. Our reward function $R$ is expressed with language output, \texttt{PASS} or \texttt{FAIL}, as a result of each hop. Naturally, we define one hop as a subtask that is completed on a specific website. For example, in the task of Fig.~\ref{fig:teaser}, the agent receives a \texttt{PASS} if it finds the correct destination in the first hop, another \texttt{PASS} for arriving at the desired flight search page at the second hop, and so on.

\paragraph{Action Space}
We follow~\citet{koh2024visualwebarena} to condense the potential agent-executed actions into a set of 12 summarized actions. Leveraging the playwright library, we simulate web pages on an X graphics server, employing a diverse array of actions to interact with the pages. These actions span a broad spectrum of behaviors mirroring human interactions with web pages, encompassing actions such as clicking on links, scrolling up and down using the scroll wheel, typing with the keyboard, etc. 
A higher hop count corresponds to a higher number of actions. On average, an MMInA task takes 12.9 actions to complete.

\paragraph{Observation Space}
The observation space $\Omega$ usually embeds partial observations of the Internet to simulate real web browsing experiences. Observations include the task descriptions and the website content. To represent web content in a reasonable length, we primarily use accessibility trees that provide a structured and unified layout of the web page content. Each node of the accessibility tree has an element ID, the type of the element, and the textual content of the element. If the element is an image, the environment downloads the image and paints the element ID on the image as a reference for the multimodal agent.

\vspace{-3mm}

\subsection{Dataset Construction}
\paragraph{Data Structure} MMInA adapted question styles from the WebQA dataset and used GPT-4V to generate similar questions with multimodal content (\textit{e.g., ``help me buy a yellow jacket online''}). The prompt structure includes: \textit{1) Instructions:} some basic concepts about tasks, accessibility tree, and actions; \textit{2) Rules:} such as ``do nothing after action \texttt{[stop]}''; \textit{3) QA examples:} some tiny examples are here to help the agent to understand instructions and above; \textit{4) Reference Websites:} this is the universe of all potential websites the agent may visit; \textit{5) Task:} a multihop multimodal task to solve. The hops vary from $1$ to $10$. Additionally, we manually crafted questions to diversify style, scope, and content across categories like shopping, search, and booking. In summary, MMInA tasks combine $2,989$ hops from $14$ dynamic, real-world websites to ensure a diverse and realistic set of challenges.

\paragraph{Annotators \& Annotation Protocols} 
Annotators are human experts in web browsing and vary in age and gender to ensure the fairness of labeling.
The annotators first proposed task templates varying in intent and difficulty. Each template generates $2-10$ tasks. All annotators then followed a ``minimalist'' approach, where annotators, acting as ``omniscient readers'', completed tasks using the shortest paths with all crucial website nodes recorded. This annotation protocol enhances the trajectory diversity in the evaluation process, where the ground-truth trajectory should be a subset of any successfully visited trajectories of the agents.

\paragraph{Performance Metrics} Evaluating real websites is challenging due to their dynamic nature, as web content frequently changes. To address this, we propose a new evaluation metric for multihop tasks based on the visited URLs. A task is considered successfully completed only when all required websites are visited in order, ensuring alignment between the agent's actions and the task objectives. Details are shown in Sec. \ref{sec:evaluation}.

\subsection{Multimodal Web Content}
Our work at MMInA focuses on multimodality-reliant tasks, which require both images and textual data to complete. For example, the task \textit{``Which one is more furry, Hi\&Yeah Comfy Faux Fur Cute Desk Chair or Armen Living Diamond Office Chair?''} requires the agent to locate and compare specified items on referenced web pages, analyzing the images and textual web page content to provide an answer.
MMInA's approach contrasts with VWA, as all tasks in our framework necessitate the processing of both visual and textual information in multiple turns (Tab.~\ref{tab:vwa_comparison}).

MMInA includes an automated process for extracting accessibility trees from web pages while identifying and downloading the image contents in the current view. This allows agents to use images alongside the accessibility tree as inputs, highlighting the critical interaction between visual and textual data in solving real-world tasks within a multimodal framework.

\subsection{Multihop Cross-website Browsing}
\begin{wrapfigure}{r}{0.2\textwidth}
\centering
\includegraphics[width=\linewidth]{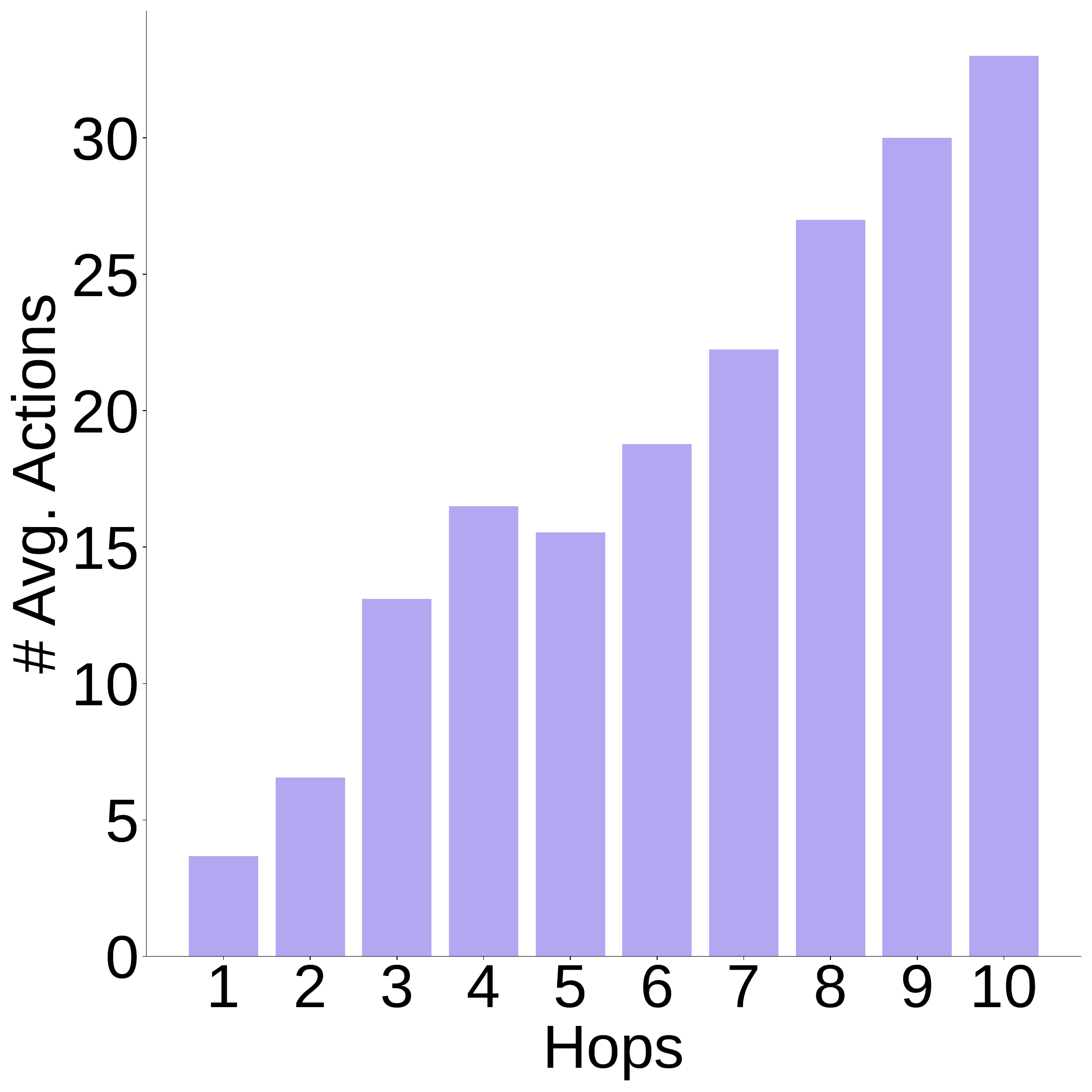}
\caption{\textbf{Average actions in multi-hop tasks.}}
\label{fig:action_dist}
\end{wrapfigure}

MMInA dataset features multihop tasks across 14 distinct websites,  covering diverse domains such as shopping, ticket booking, travel guides, and local food discovery (Fig.~\ref{fig:stats}). 
A “multihop task” requires actions across multiple websites, with the agent automatically moving to the next site after completing each hop. This setup mirrors the complexity and real-world relevance of multihop web browsing, simulating the sequence of actions a human user would typically perform when tackling a high-level task. Task descriptions include links to the available websites (see Appendix~\ref{sec:appendix_benchmark_details} for details).

\subsection{Evaluation}
\label{sec:evaluation}

\paragraph{Single-hop Evaluation}
Following~\citet{zhou2024webarena}, we implemented two evaluation methodologies for single-hop tasks to assess the semantics and effectiveness of predicted actions within the MMInA dataset. These methods offer either stringent or lenient criteria, depending on the task’s characteristics.

The first method, ``\texttt{must\_include}'', adopts a keyword-based evaluation approach. For each task, a set of essential keywords is defined. An agent's response is considered successful (\texttt{PASS}) if it incorporates all these keywords. Missing any keyword results in a failure (\texttt{FAIL}), ensuring a rigorous assessment based on keyword inclusion. The second method, ``\texttt{fuzzy\_match}'', utilizes the capabilities of advanced language models, such as GPT-3.5-Turbo, to evaluate responses. It compares the agent's response and a reference answer by asking the model: ``Given the statement \texttt{{pred}}, would it be correct to infer \texttt{{reference}}? Yes or No'', where \texttt{{pred}} is the agent's response and \texttt{{reference}} is the reference answer. The evaluation is determined by the model's reply: a \texttt{PASS} is recorded if the answer is ``Yes'', and a \texttt{FAIL} if ``No''. This method offers a flexible evaluation framework that accommodates linguistic nuances, such as assessing semantic similarities between ``gold'' and ``yellow'' in color identification tasks.
\vspace{-10pt}
\paragraph{Multihop Evaluation}
In our experiments with multihop tasks (Tab.~\ref{tab:results}), we often observe a remarkably low success rate for the entire task, if not zero. To provide a more granular evaluation, we propose an evaluation method tailored for $N$-hop problems:
The evaluation involves maintaining a queue containing the conditions of each hop's completion. In particular, the last element of a queue is always an ``END'' marker that signifies the whole multihop task is completed, making the queue's length $N+1$.  An agent succeeds at a hop once it finds out the required information (\eg, an answer string) or reaches the desired state (\eg, a specific URL). For simplicity, our benchmark enforces that the agent completes tasks in sequence, \ie, the agent is only allowed to proceed to the next hop if the current hop is correctly completed. A task is completed only if all the hops are correctly completed in sequence.

Our interleaved single-hop and multihop evaluation methods aim to provide a systematic and insightful approach to assess the performance of agents in tackling multihop tasks, addressing the challenges posed by such tasks' complexity.


\section{Experiments}

\begin{table*}[t]
\caption{\textbf{MMInA Benchmark Results.} We evaluated $4$ types of agents on the proposed MMInA benchmark: 1) LLM Agents; 2) LMM Agents; 3) Heuristic-Based Web Agents; 4) Human Baselines. Regarding to different capabilities of agents, we have different combinations of input types. Here are the definitions of all types of input: \faFileTextO: input instructions; \faCode: the accessibility tree of the current webpage; \faCc: the textual captions of the images in the current view; \faFileImageO: the images in the current view; \faBackward: the execution histories of the agent; \faTv: the original webpage. The hop success rate is defined by the percentage ($\%$) of successful visits to the targeted websites; while the task success rate is calculated by the overall percentage ($\%$) of successful tasks from the whole task set.} 
\centering
\resizebox{\textwidth}{!}{%
\begin{tabular}{@{}cccccccccccc@{}}
\toprule
\multirow{2}{*}{\textbf{Input Type}} & \multirow{2}{*}{\textbf{Agent}} & \multirow{2}{*}{\textbf{Inputs}} & \multicolumn{4}{c}{\textbf{Hop Success Rate ($\uparrow$)}} & \multicolumn{4}{c}{\textbf{Task Success Rate ($\uparrow$)}} \\
\cmidrule(lr){4-7}\cmidrule(lr){8-11}
& & & \textbf{1 hop} & \textbf{2-4 hops} & \textbf{5+ hops} & \textbf{overall} & \textbf{1 hop} & \textbf{2-4 hops} & \textbf{5+ hops} & \textbf{overall} \\
\midrule
\multirow{9}{*}{Text} 
                        & Fuyu-8B &  & 0 & 0 & 0 & 0 & 0 & 0 & 0 & 0\\
                        & CodeLLaMA-7B &  & 1.18 & 0 & 0 & 0.29 & 1.18 & 0 & 0 & 0.58\\
                        & WebShop & \multirow{1}{*}{\faFileTextO } & 20.67 & 0 & 0 & 4.17 & 20.67 & 0 & 0 & 10.12\\
                        & DeepSeek-R1-Distill-Qwen-32B & \faCode & 21.61 & 1.85 & 1.62 & 4.74 & 21.61 & 0 & 0 & 10.46\\
                        & Gemini-Pro &  & 19.09 & \textbf{34.12} & 2.13 & 11.85 &  19.09 & 0.76 & 0 & 9.54\\
                        & GPT-4 &  & 14.37 & 30.56 & 5.23 & 12.26 & 14.37 & 9.09  & 0 & 9.34\\
\cmidrule(lr){2-11}
                        & CodeLLaMA-7B  &  & 5.71  &  0  & 0 & 1.61 & 5.71 & 0 & 0 & 2.79\\
                        & WebShop  &  \multirow{1}{*}{\faFileTextO }  & 29.72 & 0.00 & 0.00 & 5.61 & 29.72 & 0 & 0 & 14.55\\
                        & DeepSeek-R1-Distill-Qwen-32B & \multirow{1}{*}{\faCode} & \textbf{47.68} & 3.84 & 4.68 & 11.11 & \textbf{47.68} & 0 & 0 & 23.07 \\
                        & Gemini-Pro  & \multirow{1}{*}{\faCc} & 30.12 & 11.09 & 0.05 & 12.38 & 30.12 & 1.52 & 0.38 & 15.22\\
                        & GPT-4  &   & 38.58 &  20.70 &  3.43 & 13.50 & 38.58 & 3.79 & 0 & 19.85\\
\midrule
\multirow{9}{*}{Multimodal} 
                        & CogAgent-9B &   & 6.92 &  0 &  0 & 1.06 & 6.92 & 0 & 0 & 3.35 \\
                        & GPT-4o & \faFileTextO & 21.90 &  9.23 &  0.96 & 5.94 & 21.90 & 3.85 & 0 & 11.61 \\
                        & Fuyu-8B & \faCode & 27.36  & 0 & 0 & 5.52 & 27.36 & 0 & 0 & 13.39\\ 
                        & Gemini-Pro-Vision & \faFileImageO & 28.94 & 16.38 & 4.03 & 10.66 &  28.94 & 1.51 & \textbf{1.13} & 18.40\\ 
                        & {GPT-4V} & & 42.91  &  21.23 &  3.99 & 13.89 & 42.91 & 3.03 & 0 & \textbf{21.77}\\
                        
\cmidrule(lr){2-11} 
                        
                        & \multirow{2}{*}{GPT-4o} & \faFileTextO & \multirow{2}{*}{27.45} & \multirow{2}{*}{17.76} & \multirow{2}{*}{\textbf{10.13}} & \multirow{2}{*}{\textbf{14.36}} & \multirow{2}{*}{27.45} & \multirow{2}{*}{3.32} & \multirow{2}{*}{0} & \multirow{2}{*}{14.04} \\  
                        & & \faCode \\
                        & \multirow{2}{*}{Gemini-Pro-Vision} & \faFileImageO  & \multirow{2}{*}{39.17} & \multirow{2}{*}{23.93} & \multirow{2}{*}{4.78} & \multirow{2}{*}{14.27} & \multirow{2}{*}{39.17} & \multirow{2}{*}{\textbf{10.61}} & \multirow{2}{*}{\textbf{1.13}} & \multirow{2}{*}{20.13}\\ 
                        & & \faBackward\\
 
\midrule
- &  Human & \faTv & 99.02 & 97.91 & 93.77 & 98.43 & 99.02 & 95.34 & 88.12 &  96.25 \\ 
\bottomrule
\end{tabular}%
}
\label{tab:results}
\end{table*}

\subsection{Baselines}
We employed a variety of state-of-the-art LLMs, LMMs, and adapted web-oriented models to evaluate their performance on the MMInA benchmark. 
For \textbf{text-based} models, we conducted evaluations in two settings:
1) \textit{text-only}: Only textual information from the website was used, ignoring image content. 2) \textit{caption-augmented}: In addition to the text, we used the BLIP-2~\cite{li2023blip} model to generate captions for website images, incorporating visual information. For \textbf{multimodal} models, both image and text information from the website were provided. More details regarding the environment, model parameters, and versions are included in Sec.~\ref{sec:appendix_experiment_details}. We categorized the models as follows: 
\paragraph{1) LLM/LMM as Agent's Reasoning Backbone}
Large language / multimodal models can act as powerful backbones in agents' reasoning processes that can predict feasible next-step actions by prompting~\cite{liu2023bolaa, zhou2024webarena}. As the textual input is the accessibility tree representation of the webpage, we categorized the text-based agents into 4 groups: 1) pretrained open-sourced LLMs, like CodeLLaMA\cite{roziere2023code} and DeepSeek-R1~\cite{guo2025deepseek}; 2) text decoders from pretrained open-sourced LMMs, like Fuyu-8b~\cite{fuyu-8b}; 3) API-based LLMs, like GPT-4~\cite{achiam2023gpt} and Gemini-Pro~\cite{team2023gemini}.
Our experiments also involved prominent LMMs such as Fuyu-8b~\cite{fuyu-8b}, Gemini-Pro-Vision~\cite{team2023gemini}, and GPT-4V~\cite{achiam2023gpt}.

\paragraph{2) Heuristic-Based Web Agents}
Several heuristic-based web agents were specifically crafted with the intention of navigation and completion of web-based tasks~\cite{yao2022webshop,deng2023mind2web,hong2024cogagent,zheng2024gpt}. We selected WebShop and CogAgent as baselines to evaluate how models trained on web-based tasks perform on the MMInA benchmark.

\paragraph{3) Human Baselines}
We conducted a comparison of hop and task performances within the same settings with an average of $3$ human test takers. The test takers come from various socioeconomic backgrounds, without information on the tasks before the evaluation. Human baselines consistently outperform all existing web agents with significant margins.

\subsection{Main Results}
The results for the different models are shown in Tab.~\ref{tab:results}, where the hop performance and task performance are evaluated respectively. 
The hop success rate reflects the percentage of successful visits to targeted websites, while the task success rate measures the percentage of tasks successfully completed by agents out of the total number of tasks.

Our experimental results indicate that current state-of-the-art models exhibit significantly reduced performance on multihop tasks. This gap highlights their struggle to effectively recognize and process structured, long-context information from web pages, which is essential for understanding web content. Additionally, performance drops sharply as the number of hops increases, revealing the models’ limitations in long-chain reasoning.

The hop success rate, which counts every successful task completion at a website, serves as an auxiliary metric to represent the procedural performance of each agent more accurately. On single-hop tasks, DeepSeek-R1-Distill-Qwen-32B outperformed all other models, indicating that the reasoning model possesses strong potential in image and context comprehension, as well as planning. However, for tasks with a hop count ranging from 2 to 4, we observed unexpected performances. Specifically, Gemini-Pro without captions and GPT-4 without captions exhibited higher performances compared to their counterparts that are augmented with captions. Further analysis of the agents' trajectories revealed that when agents were assigned relatively simple tasks while being under-informed or lacking sufficient information, they tended to ``wander'' through the given hops. This often led to an endless loop, causing them to lose focus on the task's original intent and ultimately resulting in failure.
This phenomenon explains why some agents achieve a higher hop success rate while simultaneously exhibiting a low task success rate. This insight further justifies the need for a holistic evaluation protocol with MMInA.

The experimental results showed that: 1) \textit{Multimodality-reliance}: Multimodal models exhibit overall higher performance in both hop and task performance, which makes more accurate predictions on the proposed benchmark; 2) \textit{Context window length}: Language models like CodeLLaMA and the GPT series, designed for structured and long-context processing, excel in web-based tasks relying on structured webpage representations.
Reasoning models like DeepSeek-R1 perform well in single-hop tasks due to strong contextual comprehension. However, when tackling multi-hop tasks that require retaining longer contexts, R1 struggles and exhibits degraded performance; 
3) \textit{Web-based models}: the models that were trained on web-based content (\eg, Fuyu-8B, WebShop) still exhibit the versatility and adaptability in unfamiliar environments.

\subsection{Why are Multihop Web Tasks Challenging?}

\begin{table*}[t]
    \centering
    \caption{\textbf{Evaluation by Hops.} Tables (a) and (b) display the hop success rates (SR), distinguished by hop counts (H.C.) of the tasks ranging from 2 to 6, for the models GPT-4V and Gemini-Pro-Vision, respectively. Higher success rates are marked with darker colors. Both agents fail on the early hops when solving tasks with more hops.}
    \resizebox{\textwidth}{!}{%
    \begin{minipage}{.5\linewidth}
    \resizebox{0.98\textwidth}{!}{%
        \centering
        \begin{tabular}{|c|c|c|c|c|c|c|}
            \hline
            \diagbox{H.C.}{SR(\%)} & \textbf{1st} & \textbf{2nd} & \textbf{3rd} & \textbf{4th} & \textbf{5th} & \textbf{6th} \\
            \hline
            \textbf{2} & \cellcolor[HTML]{57BB8A}56.50 & \cellcolor[HTML]{DFF2E9}11.00 & - & - & - & - \\
            \hline
            \textbf{3} & \cellcolor[HTML]{BCE4D0}22.73 & \cellcolor[HTML]{F2FAF6}4.55 & \cellcolor[HTML]{FFFFFF}0.00 & - & - & - \\
            \hline
            \textbf{4} & \cellcolor[HTML]{DAF0E6}12.50 & \cellcolor[HTML]{FFFFFF}0.00 & \cellcolor[HTML]{FFFFFF}0.00 & \cellcolor[HTML]{FFFFFF}0.00 & - & - \\
            \hline
            \textbf{5} & \cellcolor[HTML]{DBF1E6}12.28 & \cellcolor[HTML]{FAFDFC}1.75 & \cellcolor[HTML]{FFFFFF}0.00 & \cellcolor[HTML]{FFFFFF}0.00 & \cellcolor[HTML]{FFFFFF}0.00 & - \\
            \hline
            \textbf{6} & \cellcolor[HTML]{CEEBDD}16.67 & \cellcolor[HTML]{FFFFFF}0.00 & \cellcolor[HTML]{FFFFFF}0.00 & \cellcolor[HTML]{FFFFFF}0.00 & \cellcolor[HTML]{FFFFFF}0.00 & \cellcolor[HTML]{FFFFFF}0.00 \\
            \hline
        \end{tabular}
        }
        \subcaption{GPT-4V}
    \end{minipage}%
    \hfill
    \begin{minipage}{.5\linewidth}
    \resizebox{0.98\textwidth}{!}{%
        \centering
        \begin{tabular}{|c|c|c|c|c|c|c|}
            \hline
            \diagbox{H.C.}{SR(\%)} & \textbf{1st} & \textbf{2nd} & \textbf{3rd} & \textbf{4th} & \textbf{5th} & \textbf{6th} \\
            \hline
            \textbf{2} & \cellcolor[HTML]{57BB8A}69.28 & \cellcolor[HTML]{EBF7F1}8.43 & - & - & - & - \\
            \hline
            \textbf{3} & \cellcolor[HTML]{B1E0C9}32.56 & \cellcolor[HTML]{FFFFFF}0.00 & \cellcolor[HTML]{FFFFFF}0.00 & - & - & - \\
            \hline
            \textbf{4} & \cellcolor[HTML]{9FD8BC}40.00 & \cellcolor[HTML]{FFFFFF}0.00 & \cellcolor[HTML]{FFFFFF}0.00 & \cellcolor[HTML]{FFFFFF}0.00 & - & - \\
            \hline
            \textbf{5} & \cellcolor[HTML]{9AD7B9}41.67 & \cellcolor[HTML]{F3FBF7}5.00 & \cellcolor[HTML]{FFFFFF}0.00 & \cellcolor[HTML]{FFFFFF}0.00 & \cellcolor[HTML]{FFFFFF}0.00 & - \\
            \hline
            \textbf{6} & \cellcolor[HTML]{B4E1CB}31.03 & \cellcolor[HTML]{FBFEFD}1.72 & \cellcolor[HTML]{FFFFFF}0.00 & \cellcolor[HTML]{FFFFFF}0.00 & \cellcolor[HTML]{FFFFFF}0.00 & \cellcolor[HTML]{FFFFFF}0.00 \\
            \hline
        \end{tabular}
        }
        \subcaption{Gemini-Pro-Vision}
    \end{minipage}
    }
    \vspace{-6pt}
\label{tab:tri}
\vspace{-10pt}
\end{table*}

\begin{figure*}[htbp!]
    \centering
    \begin{minipage}{0.58\linewidth}
        \includegraphics[width=\linewidth]{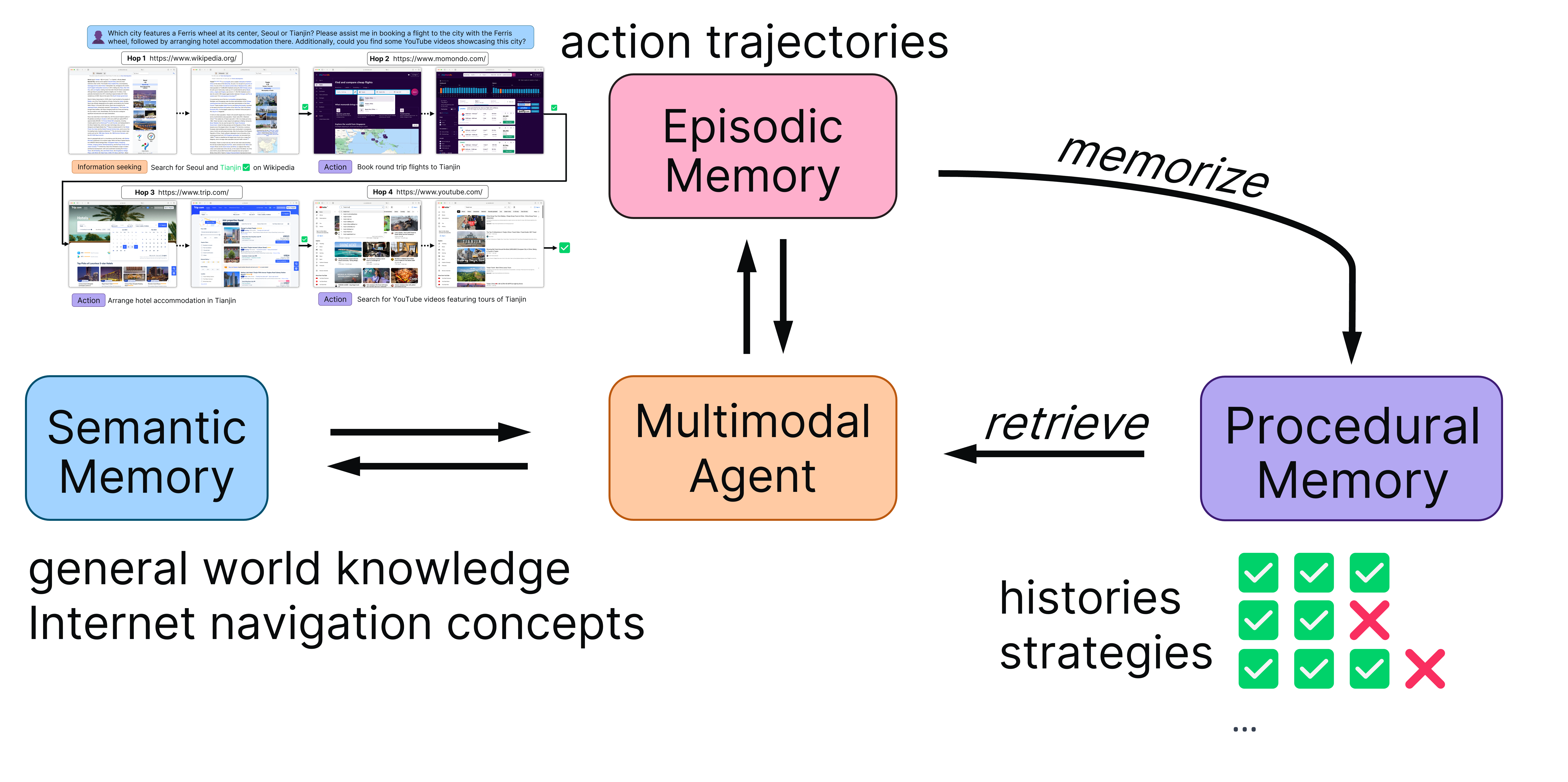}
        \caption{\textbf{Memory-augmented agents.} Our method complements LMMs by enhancing procedural memory with action trajectories on similar tasks. }
        \label{fig:agent}
    \end{minipage}
    \hfill 
    \begin{minipage}{0.41\linewidth}
        \caption{\textbf{Success rates of memory-augmented agents, by history lengths.}}
        \includegraphics[width=\linewidth]{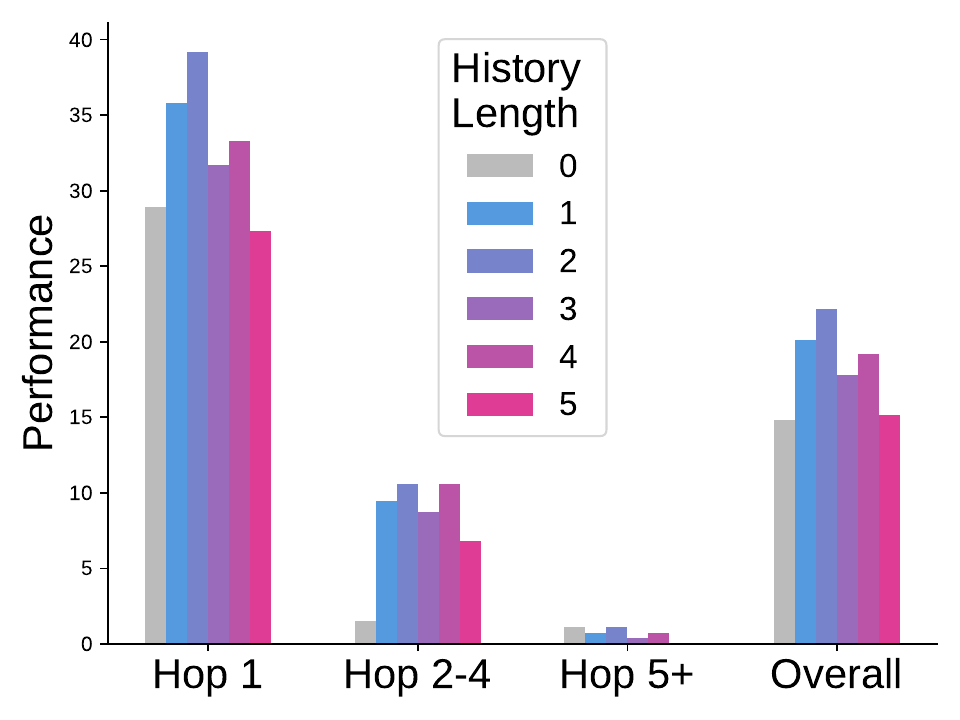}
        \label{fig:agent_ablation}
    \end{minipage}
\end{figure*}

\paragraph{Search Space}

Agents often underperform on multihop tasks, failing at early hops, yet excel when each hop is treated as an individual single-hop task.
Our analysis shows that in single-hop tasks with a single reference URL, agents tend to repeatedly attempt actions within the same website upon failure until the task is completed. Conversely, in multihop tasks where prompts include multiple websites, agents that fail on the expected site often switch to alternate sites instead of persistently retrying. This behavior leads to excessive exploration and a significant drop in task completion rates.

For example, in a complex task like ``Book a flight to Tokyo, find a tour guide, watch YouTube videos, rent a car, and book a hotel'', agents may fail in the first step. Even if they correctly redirect to a relevant site, limited memory prevents them from recalling previous steps, leading to repeated actions without progress.
Moreover, despite defined termination conditions for each hop, agents often fail to recognize and apply them, lingering in completed hops instead of advancing, ultimately failing the task.

\paragraph{Agent Input Length} The total hop count in a multihop task determines its length. Each hop's success depends on the previous one, but the success rate of any individual hop should be independent of the total number of hops, provided it remains within the specific domain.
However, after aligning the first hop semantically across all tasks, our empirical findings in Tab.~\ref{tab:tri} indicated an unexpected pattern that was contrary to the assumption above. We observed that agents performed better on tasks with fewer total hops, achieving higher success rates in completing the first hop. Conversely, as the total hop count increased, there was a noticeable decline in the success rate within the first hop. We attribute this phenomenon to the enlarged search space and the agent's weak zero-shot long-context reasoning capabilities, which we resolve in Sec.~\ref{ssec:memory-augmented}.

These findings highlight the complexity of multihop tasks, which is not simply the sum of single-hop performances but involves intricate task flow management. This complexity underlines the necessity for web agents to possess advanced planning and reasoning skills to effectively navigate and execute multihop tasks.

\subsection{Memory-augmented Agents}
\label{ssec:memory-augmented}

Agents in dynamic environments must make decisions based on real-time observations, user queries, and past trajectories. Our experiments reveal the complexity of action prediction, requiring different memory types at various stages. This highlights the need to retain information across tasks, actions, and web interactions. We propose memory-augmented web agents with three key memory systems: semantic, episodic, and procedural.
\begin{itemize}
    \vspace{-2mm}
    \item Semantic memory stores the agent’s general world knowledge, continuously updated from the Internet or knowledge bases, typically encoded in the weights of large language models.
    \vspace{-7mm}
    \item Episodic memory temporarily holds step-by-step action trajectories, enabling the agent to recall previous actions for ongoing tasks, often represented as context in autoregressive models or in-context examples.
    \vspace{-2mm}
    \item Procedural memory activates after completing a task, encoding the full action sequence and outcomes to refine strategies for future tasks.
\end{itemize}
\vspace{-5pt}
Our results emphasize the role of procedural memory in improving agent performance by replaying past action trajectories for similar tasks (see Table~\ref{tab:results}). With these memory systems, multimodal agents can access and apply relevant information, enabling more sophisticated, contextually aware responses to dynamic environments, thereby significantly enhancing performance and adaptability.

\section{Conclusion, Challenges, and Outlook}
We present MMInA, a benchmark with three key features: \textbf{1)} It benchmarks agents on real-world websites with 1,050 multimodal multihop tasks across 14 diverse websites, including experiments with state-of-the-art LLMs and LMMs, as well as human baselines; \textbf{2)} It introduces a novel holistic evaluation method for multihop tasks, assessing both task-level and hop-level success rates; \textbf{3)} It proposes a flexible memory-augmented approach to enhance agents’ performance by improving their procedural memory.
\paragraph{Future Work}
Moving forward, we will consider employing an evaluation method focused on actions, which will directly guide the agent's operations.

\section{Limitations}
Due to the protection mechanisms employed by web pages, it's exceptionally challenging to find a website that allows us to directly fetch images from HTML files. So, one of the websites we utilized is an offline standalone website, and the other is an open-source website.

\section{Ethical Considerations}
Bias in the base multimodal models can lead to inaccurate or unfair outcomes. Users should consider the representativeness of training data to avoid biased behavior.


\bibliography{custom,sections/8-reference}
\bibliographystyle{acl_natbib}

\appendix

\clearpage
\newpage

\appendix
\setcounter{table}{0}
\renewcommand{\thetable}{A\arabic{table}}
\setcounter{figure}{0}
\renewcommand{\thefigure}{A\arabic{figure}}

\section{Experiment Details}
\label{sec:appendix_experiment_details}
\subsection{Environment \& Parameter Settings}
We followed the display settings from~\cite{zhou2024webarena}, using a viewport of $1280 \times 2048$, and provided the webpage accessibility tree as text input to the models. Default parameters were used from either open-source pre-trained models or API-based models. For the web-trained agent WebShop, which was designed for a static environment and trained with specially structured queries, we used the GPT-3.5-turbo model to generate formatted queries in place of those typically sourced from the built-in environment.
For the versions of API-based models, the GPT-4 model in the paper is referred to \texttt{gpt-4-0125-preview}; the GPT-4o model is referred to \texttt{gpt-4o-2024-11-20}; the GPT-4V model is referred to \texttt{gpt-4-vision-preview}; the Gemini-Pro model is referred to \texttt{gemini-1.0-pro-001}; the Gemini-Pro-Vision is referred to \texttt{gemini-1.0-pro-vision-001}.

\subsection{Computing Resources}
\label{sec:appendix_computes}
We used 1 Nvidia RTX6000 Ada GPU with 48 GB memory for the pre-trained baseline models (most of them have 7b/8b/9b parameters) that run the inference code locally. The inference of each epoch on MMInA takes 4-8 hours, varying by model inference performance and forward times.

\subsection{Supplementary Results}

\paragraph{Hop Analysis} We follow the previous settings of hop analysis in the main paper, illustrating the agent performance of a GPT-4V agent in Tab.~\ref{tab:tri_gpt}. We observed again that agents performed better on tasks with fewer total hops, achieving higher success rates in completing the first hop. Conversely, as the total hop count increased, there was a noticeable decline in the success rate within the first hop. Because there are fewer long-range (>7 hops) tasks, the success rates fluctuate due to randomness.

\paragraph{Ablation of agents' memory} We enhance LMMs with memory by appending action trajectories from the last $K$ tasks, comprising task descriptions and web content observations, to their prompts. This approach, which integrates replayed experiences, helps narrow the agent's search space, thereby grounding its reasoning. However, this technique multiplies the input length by $K$, presenting a challenge for LMMs accustomed to shorter inputs. To find a balance, we determined the optimal $K$ value for constructing procedural memory, which, as illustrated in 
is typically $K=2$. Our tests show that agents with procedural memory enhancements perform better in action prediction and execution. Yet, we observed a non-linear relationship between the number of historical references and performance. In simpler tasks within domains like shopping or Wikipedia, a smaller historical set—specifically $K=1$ or $2$—tended to yield better results, while larger histories offered diminishing returns and introduced biases and disturbances into the decision-making process. Although these experiments were conducted using Gemini-Pro-Vision, our method is designed to be model-agnostic, adaptable to any LMM or LLM.

\begin{table*}[h]
\caption{\textbf{Performance of GPT-4V on multihop tasks based on the hop number ranging from 2 to 10.} The success rate (sr) is calculated based on single-hop evaluation results over the whole completed number of hops.}
\centering
\begin{tabular}{@{}ccrrrrrrrrrr@{}}
\toprule
 GPT4V & count & sr1   & sr2  & sr3  & sr4 & sr5 & sr6 & sr7 & sr8 & sr9 & sr10 \\ \midrule
2-h & 200   & 56.50 & 11.00 &      &     &     &     &     &     &     &      \\
3-h & 44    & 22.73 & 4.55  & 0.00 &     &     &     &     &     &     &      \\
4-h & 16    & 12.50 & 0.00  & 0.00 & 0.00&     &     &     &     &     &      \\
5-h & 57    & 12.28 & 1.75  & 0.00 & 0.00& 0.00&     &     &     &     &      \\
6-h & 60    & 16.67 & 0.00  & 0.00 & 0.00& 0.00& 0.00&     &     &     &      \\
7-h & 59    & 25.42 & 0.00  & 0.00 & 0.00& 0.00& 0.00& 0.00&     &     &      \\
8-h & 35    & 40.00 & 0.00  & 0.00 & 0.00& 0.00& 0.00& 0.00& 0.00&     &      \\
9-h & 30    & 56.67 & 20.00 & 3.33 & 0.00& 0.00& 0.00& 0.00& 0.00& 0.00&      \\
10-h & 19    & 52.63 & 0.00 & 0.00 & 0.00 & 0.00 & 0.00 & 0.00 & 0.00 & 0.00 & 0.00 \\ \bottomrule
\end{tabular}
\vspace{0.1in}
\label{tab:tri_gpt}
\end{table*}

\section{MMInA Benchmark Details}
\label{sec:appendix_benchmark_details}

\subsection{Datasets}
\subsubsection{Annotators}
MMInA is constructed by three human annotators from scratch. They, varying in age and gender but proficient in web browsing, were provided pre-annotated examples and followed consistent guidelines. Each labeled different dataset portions, and cross-validation was conducted for task diversity and answer accuracy. All annotators signed formal agreements and were trained for annotation.
\subsubsection{Anontation Protocals}
The final trajectory includes all necessary website nodes. Agents, however, can freely explore and visit unnecessary websites before completing tasks.
\subsubsection{Data Statistics}
From the hop counts shown in Fig.~\ref{fig:action_dist}, we observe that as the number of hops increases, the number of actions required by the agent also increases. 
However, it's worth noting that the average number of actions required for 5-hop data is lower than that for 4-hop data. In our dataset, 4-hop content involves comparative operations, such as \textit{``Which one has a Ferris wheel in the center of the city, Tianjin or Chengdu?''} Since our definition of multi-hop tasks involves navigation across different web pages, we do not categorize these comparative questions as 5-hop tasks. 

Most of our multitasking revolves around travel. First, we need to determine the travel destination. We let the agent determine the travel destination by retrieving and answering questions on Wikipedia. A set of tasks is related to travel, including booking flights, reserving hotels, exchanging currencies, etc. Then we randomly select some tasks from the task set and combine them into a complete and smooth task. In each task's JSON file, we include keywords such as ``flight'', which is the three-letter code of the destination airport. The reason for these keywords is that for specific tasks, some sub-tasks cannot be measured with a unified standard, so we add these keywords to judge the endpoints of specific tasks. Another type of task related to cooking includes purchasing food on Amazon and searching for recipes. Each task is close to real life and uses real and dynamic web pages for operation. We have 1050 tasks in total and each task contains a QA pair and other supportive materials. 108 QA pairs are filtered from WebQA~\cite{chang2022webqa}. Statistical details about our dataset are shown in Fig.~\ref{fig:stats}.

\subsection{Tasks}

\subsubsection{Website Links}
Tab.~\ref{tab:websites} reveals the links to each website defined in MMInA.
\begin{table*}[htbp!]
    \centering
    \begin{threeparttable}
    \small
    \caption{\textbf{Links to MMInA websites.} All of them are evolving real-world websites with content updated over time, except Shopping. Our flexible evaluation protocol facilitates supporting more websites in the future.}
    \begin{tabular}{@{}p{4cm}p{8cm}@{}}
    \hline
    \textbf{Description} & \textbf{URL} \\ 
    \hline
    Wikipedia\tnote{1} & \url{https://library.kiwix.org/viewer#wikipedia_en_all_maxi_2024-01/A/User%3AThe_other_Kiwix_guy/Landing} \\
    Car renting & \url{https://www.trip.com/carhire/} \\
    Flight booking & \url{https://www.momondo.com/} \\
    Hotel booking & \url{https://www.trip.com/hotels/} \\
    Event searching & \url{https://www.eventbrite.com/} \\
    Twitter & \url{https://twitter.com/home} \\
    Amazon & \url{https://www.amazon.com/} \\
    YouTube & \url{https://www.youtube.com/} \\
    Find food & \url{https://www.timeout.com/} \\
    Exchange dollars & \url{https://www.xe.com/} \\
    Travel guide & \url{https://www.nomadicmatt.com} \\
    Recipes & \url{https://www.allrecipes.com/}\\
    Train booking & \url{https://www.trip.com/trains/}\\
    Shopping & OneStopMarket (an offline standalone website) \\
    \hline
    \end{tabular}
    \label{tab:websites}
    \begin{tablenotes}
        \footnotesize
        \item[1] Since libraries in Kiwix may update, resulting in URLs with advanced dates, it's advisable to verify the Wikipedia library on the official Kiwix page. However, this doesn't affect our experiments.
      \end{tablenotes}
    \end{threeparttable}
\end{table*}

\subsubsection{Comparision with other benchmarks}
The comparison of our benchmark and the VWA benchmark is shown in~\ref{tab:vwa_comparison}.
\begin{table}[htbp!]
\centering
\caption{\textbf{Comparison between VWA~\cite{koh2024visualwebarena} and MMInA.} MMInA requires multimodal inputs at multiple steps to accomplish the task, which makes it a more challenging multimodal benchmark.}
\begin{tabular}{m{0.4\linewidth}|m{0.4\linewidth}}
\hline
\textbf{Websites} & \textbf{Task} \\
\hline
\multicolumn{2}{l}{VWA} \\
\hline
\includegraphics[width=\linewidth]{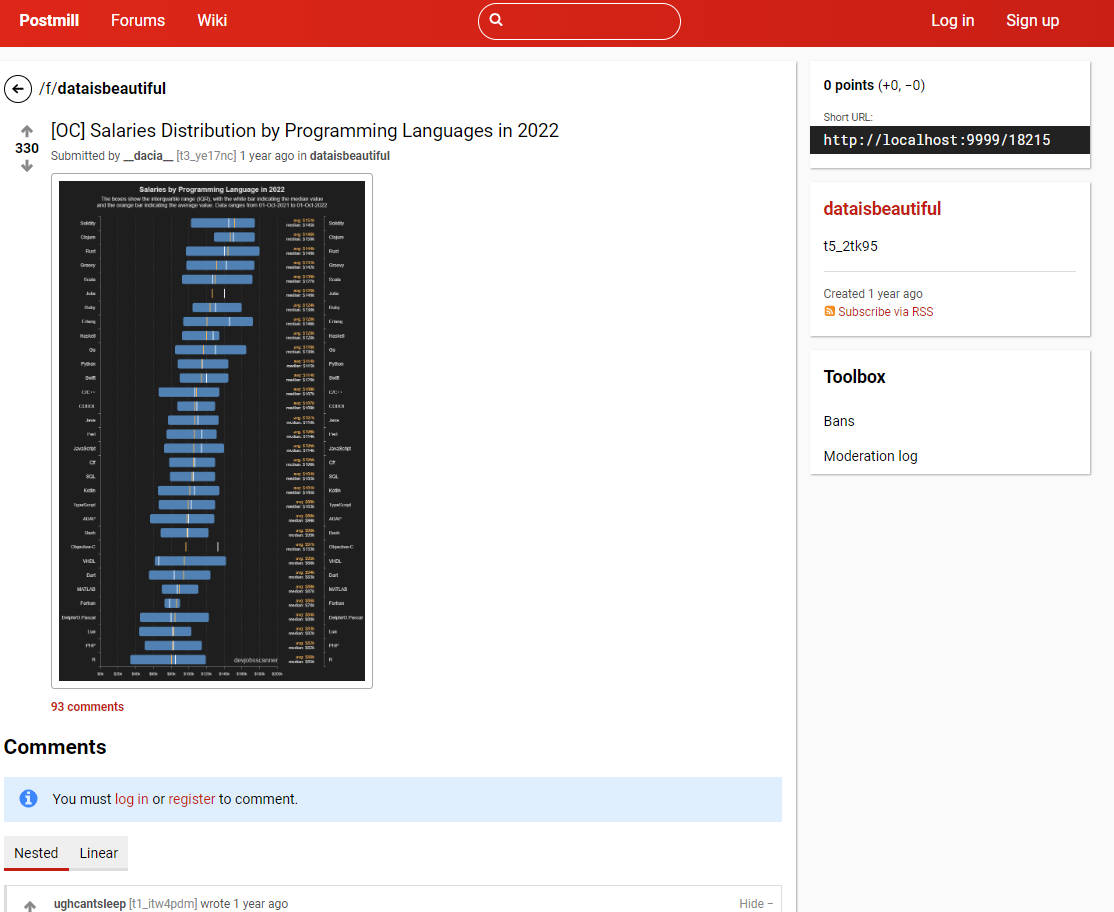} & \textit{When did the programming language that has the largest variance in salary first appear? Answer using the information from the Wikipedia site in the second tab.} \\
\hline
\multicolumn{2}{l}{MMInA} \\
\hline
\includegraphics[width=\linewidth]{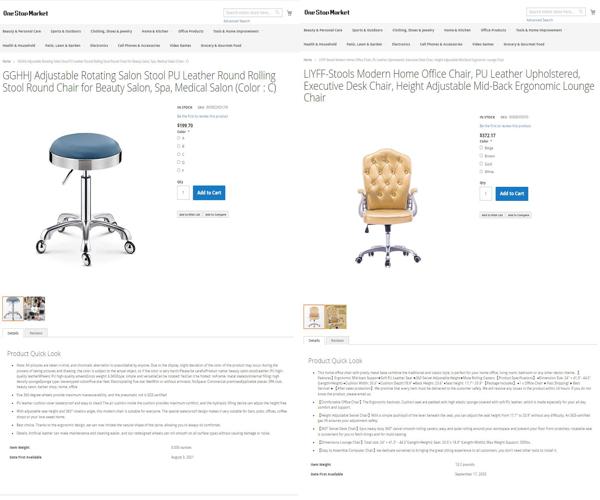} & \textit{Do both LIYFF-Stools Modern Home Office Chair and GGHHJ Adjustable Rotating Salon Stool PU Leather Round Rolling Stool Round Chair have armrests?} \\
\hline
\end{tabular}
\label{tab:vwa_comparison}
\vspace{-5mm}
\end{table}

\section{More Related Works}
\label{sec:appendix_related workds}

The literature review will be divided into several key sections, each focusing on a critical aspect of research related to the development of multimodal autonomous web agents. This comprehensive overview will delve into multimodal datasets, large language/multimodal models as backbones, and various types of autonomous agents, including embodied agents and web agents. The aim is to provide a thorough understanding of the current state of research in these areas, as well as to identify gaps and provide insights for future work.
\paragraph{Multimodal Datasets}
Recent progress in multimodal learning, showcased by models like CLIP\cite{radford2021learning}, DALL-E\cite{ramesh2021zero}, Stable Diffusion\cite{rombach2022high}, Flamingo\cite{alayrac2022flamingo}, and GPT series, has led to significant improvements in areas such as zero-shot classification, image creation, and in-context learning. Although these models employ various algorithmic approaches, including contrastive learning, diffusion techniques, and auto-regressive models, they share a fundamental reliance on large datasets comprising image-text pairs. This commonality underscores the importance of such datasets in driving advancements in multimodal AI capabilities. 

Webdataset\footnote{\url{https://github.com/webdataset/webdataset}} is a commonly used dataset as it contains thousands of image-text pairs data scraped from websites; LAION-5B\cite{schuhmann2022laion} is a dataset that contains 5.85 billion CLIP-filtered image-text pairs, where 2.32B contain English language; MIMIC-IT\cite{li2025otter} is a dataset consists of 2.8 million multimodal instruction-response pairs equipped with rich in-context information, with 2.2 million unique instructions derived from images and videos; DataComp\cite{gadre2023datacomp} is a newly brought dataset which consists of four stages: A) Deciding on a scale that fits within resource limitations. B) Creating a dataset, opting for either the filtering approach or the Bring Your Own Data (BYOD) track. C) Using a set architecture and specific hyperparameters to train a CLIP model on the created dataset. D) Assessing the performance of the trained model across a variety of downstream tasks.

\paragraph{Large Language/Multimodal Models as Backbones}
Instruction tuning is a common method used in LLM training, which involves refining pre-trained LLMs using datasets formatted as instructions. This approach enhances the model's ability to perform new, unseen tasks by simply following directions, thereby improving its zero-shot capabilities. Some notable models like ChatGPT\cite{achiam2023gpt}, InstructGPT\cite{ouyang2022training}, FLAN\cite{wei2021finetuned, chung2024scaling} are built on top of instruction tuning methods.

Inherited the success from LLMs, LMM training is also extended to the instruction-tuning methods by utilizing the multimodal instruction data, which contains: a textual \texttt{<instruction>} to describe the task; \texttt{{<image>, <text>}} pair as input to enable the multimodalities; the model output with a token \texttt{<output><EOS>} to identify the end of the output. A multimodal instruction sample can be denoted in a triplet form, i.e., $I, M, R$, where \( I, M, R \) represent the instruction, the multimodal input, and the ground truth response, respectively. The LMM predicts an answer given the instruction and the multimodal input:
\[ A = f(I, M; \theta) \] the optimizing objective can be formulated as:
\begin{equation}
    \mathcal{L}(\theta) = - \sum_{i=1}^{N} \log p(R_i | I, R_{<i};\theta)
\end{equation}

\textbf{1) Transformation} The efficacy of instruction tuning in the training of LMMs is significantly constrained by the limitations in length and type of data available in current Visual Question Answering (VQA) datasets. To address this, some researchers have opted to adapt the provided instructions, transforming the succinct answer data into extended sentences enriched with semantic details~\cite{zhao2023chatbridge}. Other studies, such as in, reconstructed the answer by prompting ChatGPT to emulate the capabilities of advanced language models.

\textbf{2) Self-Instruct} LLaVA \cite{liu2024visual} extends the multimodal approach by converting images into descriptive texts and outlines of bounding boxes, then uses GPT-4 to create additional data within the context provided by initial examples.

\paragraph{Autonomous Agents in Virtual World}
Agents designed for Graphical User Interfaces (GUIs) are crafted to streamline complex activities on digital devices like smartphones and desktops. These GUI agents may employ HTML as inputs or use screenshots to facilitate task execution in a broader context.
Traditionally, research has revolved around training these agents in restrictive, static environments, a practice that deviates from human learning and hinders the agents' ability to make decisions akin to humans. However, the emergence of large language models (LLMs)  and large multimodal models (LMMs) equipped with vast web knowledge marks a pivotal shift towards achieving a more human-like intellect in agents, sparking a surge in research on LLM/LMM-enhanced autonomous agents. This section aims to explore the latest state-of-the-art (SOTA) developments in autonomous agents, examining both web GUI agents and mobile GUI agents.

\textbf{1) GUI Agents - Web Agents} Despite the current progress of web agents discussed in the main paper, several works also explored the development of web agents.
TravelPlanner~\cite{xie2024travelplanner} proposes a benchmark that provides a sandbox environment with tools for accessing nearly four million data records. It includes 1,225 planning intents and reference plans to evaluate the planning strategies of language agents by using tools;
OmniACT~\cite{kapoor2024omniact} presents a dataset and benchmark for assessing an agent’s capability to generate executable programs for computer tasks. It uses the PyAutoGUI Python library to automate mouse and keyboard operations across different operating systems and web domains. It addresses the limitations of HTML-based agents by providing a multimodal challenge where visual cues are crucial, thus enabling a more robust understanding of UI elements, but it still shows the inability to handle native desktop applications or multi-application tasks;
WEBLINX~\cite{lu2024weblinx} also proposes a benchmark for conversational web navigation, addressing the problem of enabling a digital agent to control a web browser and follow user instructions in a multi-turn dialogue fashion. The method involves a retrieval-inspired model that prunes HTML pages by ranking relevant elements, addressing the issue of LLMs not being able to process entire web pages in real-time. The technology used includes a dense markup ranker for element selection and multimodal models that combine screenshots, action history, and textual website representation. The performance is evaluated on tasks like creating a task on Google Calendar, with the model’s ability to replicate human behavior when navigating the web;
DUAL-VCR~\cite{kil2024dual} leverages the “dual view” of HTML elements in webpage screenshots, contextualizing each element with its visual neighbors. This approach uses both textual and visual features to create more informative representations for decision-making.

\textbf{2) GUI Agents - Mobile Agents} Besides web agents, mobile GUI agents are gaining more and more popularity, which are developed to handle intricate tasks automatically on digital devices like smartphones.
ERICA~\cite{deka2016erica} defines a system for interaction mining in Android applications. It employs a human-computer interaction approach to capture the data, making it scalable and capable of capturing a wide range of interactions. PIXEL~\cite{rust2023language} and Pix2Struct~\cite{lee2023pix2struct} show promising capability in multilingual transfer and UI navigation, respectively, but they struggle with language understanding tasks compared to text-only LMs like BERT, limiting their utility. Patch-and-Text Prediction (PTP) proposed in~\cite{gao2024improving} leads to better language understanding capabilities by masking and recovering both image patches and text within screenshots; 
AppAgent~\cite{zhang2025appagent} presents a multimodal agent that operates smartphone apps through low-level actions like tapping and swiping, mimicking human interactions. The agent learns app functionalities through exploration, either autonomously or by observing human demonstrations, and then applies this knowledge to execute tasks;
Mobile-Agent~\cite{wang2024mobile}, which uses visual perception tools to locate operations on a mobile device using screenshots. It involves OCR models for identifying visual and textual elements, while realizing self-planning and self-reflection to autonomously navigate mobile apps, with a benchmark called Mobile-Eval introduced for performance evaluation;
SeeClick~\cite{cheng2024seeclick} is a visual GUI agent that operates solely on screenshots, bypassing the need for structured text. It employs Large Vision-Language Models (LVLMs) enhanced with GUI grounding pre-training to accurately locate screen elements based on instructions. The method involves automating the curation of GUI grounding data and creating a GUI grounding benchmark named ScreenSpot. It adapts universally to various GUI platforms and relies on screenshots. Simplifying the action space to clicking and typing.

\subsection{Other Benchmarks}
\citet{shi2017world} and \citet{liu2018reinforcement} establish a platform of website widgets where agents can complete online tasks through basic keyboard and mouse operations. 
APIBench~\cite{li2023api}, introduced by Gorilla\cite{patil2024gorilla}, is a tool-augmented LLM benchmark to assess the tool utilization abilities of agents for code generation tasks. AgentBench~\cite{liu2024agentbench} steps forward to provide a more general toolkit with lots of closed-box environments to assess agents' performances in answering user queries. 

Webshop~\cite{yao2022webshop} builds a simulated e-commerce environment featuring 1.18 million real-world products, complemented by 12,087 crowdsourced textual instructions. It converted the action prediction task into a choice-based imitation learning process, which facilitated the accuracy and ability for task execution. However, this approach failed to evaluate open-ended agent actions in the real world. It was also limited by its monotonous design of a one-website environment, which resulted in only a single category of web browsing tasks. Mind2Web~\cite{deng2023mind2web} tries to construct a generalist web agent, which creates a dataset for crafting and benchmarking web agents by the ability of instruction following. It proposed a two-stage training to convert the action prediction problem into MCQs. SeeAct~\cite{zheng2024gpt} is a following work that enabled multimodal information for visually understanding rendered webpages and generating more accurate action plans. WebVoyager~\cite{he2024webvoyager} is capable of capturing screenshots of web pages and then using JavaScript tools to automatically identify interactive elements based on the types of webpage elements. WebArena~\cite{zhou2024webarena} deploys a standalone set of multicategory websites in an interactive environment. VisualWebArena~\cite{koh2024visualwebarena} is a subsequent project that built upon the foundation of WebArena, introducing the reliance on visual cues into the benchmark's design. The tasks of existing benchmarks are oversimplified whose completions requiring a single website, which is highly diverged from the natural web browsing tasks and should originally be designed for multihop over a long-horizon setting.

\end{document}